\documentclass[11pt,reqno]{amsart}
\usepackage{amsmath,amssymb,amsthm}
\usepackage{graphicx,epsfig,epstopdf}
\usepackage{colortbl,framed}

\renewcommand{\qed}{\hfill{\tiny \ensuremath{\blacksquare} }}%

\renewcommand{\qed}{\hfill {\tiny {\ensuremath{\blacksquare}}}}

\newtheorem{theorem}{Theorem}[section]

\newtheorem{lemma}{Lemma}[section]

\newtheorem{assumption}{Assumption}[section]

\theoremstyle{definition}

\newtheorem{remark}{Comment}[section]
\numberwithin{remark}{section}

\numberwithin{equation}{section}
\numberwithin{theorem}{section}

\usepackage{url}
\usepackage[dcucite]{harvard}
\newcommand{\citen}{\citeasnoun}

\newcommand{\Ep}{{\mathrm{E}}}

\renewcommand{\Pr}{{\mathrm{P}}}

\renewcommand{\hat}{\widehat}

\renewcommand{\Pr}{{\mathrm{P}}}

\renewcommand{\hat}{\widehat}
\renewcommand{\leq}{\leqslant}
\renewcommand{\geq}{\geqslant}

\renewcommand{\[}{\left[}

\begin{document}
\begin{center}
\bigskip
{\Large Double/Debiased/Neyman Machine Learning of Treatment Effects}\\

 by
 
\authors{\textsc{Victor Chernozhukov, Denis Chetverikov, Mert Demirer,
Esther Duflo, Christian Hansen, and Whitney Newey}}\thanks{This is a conference version 
of  \citen{CCDHM16}.   To appear in: American Economic Review 2017 (May), Papers \& Proceedings. }

\end{center}

\title[Double Machine Learning]{}
\date{January, 2017.}
\maketitle

\begin{footnotesize}
\textbf{Abstract.} \citen{CCDHM16} provide a generic double/de-biased machine learning (DML) approach for obtaining valid inferential statements about focal parameters, using Neyman-orthogonal scores and cross-fitting, in settings where nuisance parameters are estimated using a new generation of nonparametric fitting methods for high-dimensional data, called machine learning methods.  In this note, we illustrate the application of this method in the context of estimating average treatment effects (ATE) and average treatment effects on the treated (ATTE) using observational data.  A more general discussion and references to the existing literature are available in \citeasnoun{CCDHM16}.


\end{footnotesize}

\section{Scores for Average Treatment Effects}

We consider estimation of ATE and ATTE under the unconfoundedness assumption of \citeasnoun{RR:prop}.  We consider the case where treatment effects are fully heterogeneous and the treatment variable, $D$, is binary, $D \in \{0,1\}$.  We let $Y$ denote the outcome variable of interest and $Z$ denote a set of control variables.  We then model random vector $(Y,D,Z)$ as
\begin{eqnarray}\label{eq: HetPL1}
& Y  = g_0(D, Z) + \zeta,  &  \text{E}[\zeta \mid Z, D]= 0,\\
\label{eq: HetPL2}
& D  = m_0(Z) + \nu,   &   \text{E}[\nu\mid Z] = 0.
\end{eqnarray}
Since $D$ is not additively separable, this model allows for very general heterogeneity in treatment effects.  Common target parameters $\theta_0$ in this model are the ATE,
$$
\theta_0 = \text{E}[ g_0(1,Z) - g_0(0,Z)],
$$ 
and the ATTE,
$$
\theta_0 = \text{E}[ g_0(1,Z) - g_0(0,Z)|D=1].
$$

The confounding factors $Z$ affect the treatment variable $D$ via the propensity score, $m_0(Z) := \text{E}[D|Z],$ 
and the outcome variable via the function $g_0(D, Z)$. Both of these functions are unknown and potentially complicated, and we consider estimating these functions via the use of ML methods.

We proceed to set up moment conditions with scores that obey a type of orthogonality with respect to nuisance functions. Specifically, we make use of scores $\psi (W; \theta, \eta)$ that satisfy the identification condition
\begin{align}\label{eq: moment}
\text{E} \psi(W; \theta_0, \eta_0) = 0,
\end{align}
and the \textit{Neyman orthogonality condition}
\begin{align}\label{eq: orthogonality}
\partial_\eta  \text{E} \psi (W; \theta_0, \eta) \Big |_{\eta = \eta_0} =  0,
\end{align}
where $W = (Y,D,Z)$, $\theta_0$ is the parameter of interest,  $\eta$ denotes nuisance functions with population value $\eta_0$,  $\partial_\eta f  \ |_{\eta = \eta_0} $ denote the derivative  of $f$ with respect
to $\eta$ (the Gateaux derivative operator). 

Using moment conditions that satisfy (\ref{eq: orthogonality}) to construct estimators and inference procedures that are robust to small mistakes in nuisance parameters has a long history in statistics, following foundational work of \citeasnoun{Neyman59}. 
Using moment conditions that satisfy (\ref{eq: orthogonality}) is also crucial to developing valid inference procedures for $\theta_0$ after using ML methods to produce estimators $\hat \eta_0$ as discussed, e.g., in \citeasnoun{CHS:AnnRev}. 
 In practice, estimation of $\theta_0$ will be based on the empirical analog of (\ref{eq: moment}) with $\eta_0$ replaced by $\hat\eta_0$, and the Neyman orthogonality condition (\ref{eq: orthogonality}) ensures sufficient insensitivity to this replacement that high-quality inference for $\theta_0$ may be obtained.  The second \textit{critical ingredient}, that enables the use of  wide array of modern ML estimators is  \textit{data splitting}, as discussed in the next section.  


Neyman-orthogonal scores are readily available for both the ATE and ATTE -- one can employ the doubly robust/efficient scores of \citeasnoun{robins:dr} and \citeasnoun{hahn:prop}, which are automatically Neyman orthogonal. For estimating the ATE, we employ 
 \begin{align}\begin{split}\label{ATE-setup}
 \psi(W;  \theta,  \eta) & :=    (g(1, Z) - g(0,Z)) + \frac{D (Y -g(1,Z)) } {m(Z)} - \frac{(1-D) (Y -g(0,Z))}{1-m(Z)}  -\theta, \\
 \eta (Z)  & := (g(0, Z), g(1, Z), m(Z) ), \quad 
 \eta_0(Z)  :=  (g_0(0, Z), g_0(1, Z), m_0(Z) ),\end{split}\end{align}
where  $\eta (Z)$ is the nuisance parameter with true value denoted by $\eta_0(Z)$ consisting of $P$-square integrable functions, for $P$ defined in Assumption \ref{ass:ATE}, 
mapping the support of $Z$ to $\mathbb{R}\times\mathbb{R}\times (\varepsilon,1-\varepsilon)$ where $\varepsilon>0$ is a constant.
For estimation of ATTE, we use the score
 \begin{align}\begin{split}\label{ATTE-setup}
& \psi(W; \theta , \eta) =   \frac{D (Y -g(0,Z) ) } {m} - \frac{m(Z)(1-D) (Y -g(0,Z))}{(1-m(Z))m} -\theta \frac{D}{m }, \\
\eta (Z)  & := (g(0, Z), g(1, Z), m(Z), m),\quad \eta_0(Z) = (g_0(0, Z), g_0(1, Z), m_0(Z), \Ep[D] ),\end{split}\end{align}
where again $\eta (Z)$ is the nuisance parameter with true value denoted by $\eta_0(Z)$ consisting of three $P$-square integrable 
functions, for $P$ defined in Assumption \ref{ass:ATE},  mapping the support of $Z$ to $\mathbb{R}\times\mathbb{R}\times (\varepsilon,1-\varepsilon)$
and a constant $m \in (\varepsilon,1-\varepsilon)$.
The respective scores for ATE and ATTE obey
the identification condition (\ref{eq: moment}) and the Neyman orthogonality property
(\ref{eq: orthogonality}).  Note that all semi-parametrically efficient scores share the orthogonality property (\ref{eq: orthogonality}), but not all orthogonal scores are efficient. In some problems, we may use inefficient orthogonal scores to have more robustness.    Moreover, the use of efficient scores could be considerably refined using the targeted maximum likelihood approach of \citeasnoun{srr:rejoinder} and \citeasnoun{vdl:rubin} in many contexts.

\section{Algorithm and Result}\label{Sec: Algorithm}

We describe the DML estimator of $\theta_0$ using random sample $(W_i)_{i=1}^N$.  The algorithm makes use of a form of sample splitting, which we call cross-fitting. It builds on the ideas e.g. in \citeasnoun{ssiv}.  The use of sample-splitting is a \textit{crucial ingredient} to the approach that helps avoid \textit{overfitting} which can easily result from the application of complex, flexible methods such as boosted linear and tree models, random forests, and various ensemble and hybrid ML methods.   While the use of the Neyman orthogonality serves to reduce regularization and modeling biases of  ML estimators $\hat \eta_0$, the data splitting serves to eliminate the overfitting bias, which may easily occur, with the data scientist being even unaware of it.  Thus the DML method employs the \textit{double debiasing}.

\smallskip

\begin{framed}
\noindent\textbf{Algorithm: Estimation using Orthogonal Scores by $K$-fold Cross-Fitting}

\textbf{Step 1.} Let $K$ be a fixed integer. Form a $K$-fold random partition of  $\{1,..., N\}$ by dividing it into equal parts $(I_k)_{k=1}^K$ each of size $n:=N/K$, assuming that $N$ is a multiple of $K$.  For each set $I_k$, let $I_k^c$ denote all observation indices that are not in $I_k$.

\textbf{Step 2.} Construct $K$ estimators 
$$
\check \theta_{0}(I_k, I_k^c), \quad k=1,...,K,
$$  
that employ the machine learning estimators 
$$\hat \eta_{0}(I^c_k) =  \left (\hat g_0(0, Z; I^c_k), \ \hat g_0(1, Z; I^c_k),\  \hat m_0(Z; I^c_k), \ \frac{1}{N - n} \sum_{i\in I_k^c} D_i \right )',$$
of the nuisance parameters  
$$
\eta_0(Z) = (g_0(0, Z), g_0(1, Z), m_0(Z), \text{E}[D])',
$$ 
and where each estimator $\check \theta_0(I_k,I_k^c)$ 
is defined as the root $\theta$ of 
$$
\frac{1}{n} \sum_{i \in I_k} \psi(W; \theta, \hat \eta_0(I^c_k)) =0,
$$
for the score $\psi$ defined in (\ref{ATE-setup}) for the ATE and in (\ref{ATTE-setup}) for the ATTE. 

\textbf{Step 3.} Average the $K$ estimators to obtain the final estimator: 
\begin{align}\label{eq:aggregatemore}
\tilde \theta_0 = \frac{1}{K} \sum_{k=1}^K \check \theta_{0}(I_k,I^c_k).
\end{align}
An approximate standard error for this estimator is  $\hat \sigma/\sqrt{N}$, where
$$
\hat \sigma^2 = \frac{1}{N} \sum_{i=1}^N \hat \psi_i^2,
$$
$\hat \psi_i := \psi(W_i; \tilde \theta_0, \hat \eta_0(I^c_{k(i)}))$, and $k(i) := \{ k \in \{1,...,K\}: i \in I_k\}$.  
An approximate $(1-\alpha)\times 100\%$ confidence interval is 
$$
\text{CI}_n := [ \tilde \theta_0 \pm \Phi^{-1} (1-\alpha/2)  \hat \sigma /\sqrt{N}].
$$
\end{framed}

We now state a formal result that provides the asymptotic properties of $\tilde\theta_0$.
Let $(\delta_n)_{n=1}^\infty$ and $(\Delta_n)_{n=1}^\infty$
be sequences of positive constants approaching 0.  Let $c, \varepsilon, C$ and $q>4$ be fixed positive constants, and let $K$ be a fixed integer.

\begin{assumption}\label{ass:ATE}  Let $\mathcal{P}$ be the set of probability distributions $P$ for $(Y,D,Z)$ such that (i)  equations (\ref{eq: HetPL1})-(\ref{eq: HetPL2}) hold, with $D \in \{0,1\}$, (ii)  the following conditions on moments hold for all $d \in \{0,1\}$: $\|g(d,Z)\|_{P,q} \leq C$, $\|Y\|_{P,q} \leq C$, $P( \varepsilon \leq m_0(Z) \leq 1- \varepsilon) =1$, $P(\Ep_P[\zeta^2\mid Z] \leq C) = 1$, $\|\zeta\|_{P,2} \geq c$, and $\|\nu\|_{P,2}\geq c$ and (iii) the ML estimators of the nuisance parameters based upon a random subset $I^c_k$ of $\{1,..., N\}$ of size $N-n$, obey the following conditions for all $N \geq 2K$ and $d \in \{0,1\}$:  $\| \hat g_0(d,Z; I^c_k) - g_0(d,Z) \|_{P,2} \cdot  \| \hat m_0(Z;I^c_k) - m_0(Z) \|_{P,2} \leq \delta_n n^{-1/2}$, $\| \hat g_0(d,Z; I^c_k) - g_0(d,Z) \|_{P,2} +  \| \hat m_0(Z;I^c_k) - m_0(Z) \|_{P,2} \leq \delta_n$, and  $P(\varepsilon \leq \hat m_0(Z;I^c_k) \leq 1- \varepsilon) =1$, with $\Pr_P$-probability  no less than $1- \Delta_n$.\end{assumption}

The assumption on the rate of estimating the nuisance
parameters is a non-primitive condition.  These rates of convergence are available for most often used ML methods and are case-specific, so we do not restate conditions that are needed to reach these rates.  

\begin{remark}[Tighness of conditions] The conditions are fairly sharp (though somewhat simplified for presentation sake).   The sharpness can be understood by examining the case where 
regression function $g_0 $ and propensity function $m_0$ are sparse with sparsity indices $s^g \ll n$ and $s^m \ll n$, and estimators $\hat g_0 $ and $\hat m_0$ having sparsity indices of orders $s^g$ and $s^m$ and covering to $g_0$  and $m_0$ at the
rates $\sqrt{{s^g}/{n}}$ and $\sqrt{{s^m}/{n}}$, ignoring logs, based on $\ell_1$-penalized estimators.  Then the rate conditions in the assumption require 
$$ \sqrt{s^g/n} \sqrt{s^m/n} \ll n^{-1/2} \Leftrightarrow s^g s^m \ll \sqrt n  $$ (ignoring logs) which is much weaker than the condition $(s^g)^2 + (s^m)^2 \ll n$ (ignoring logs) terms, required without sample splitting.  For example, if the propensity score $m_0$ is very sparse, then the regression function is allowed to be quite dense, and vice versa.  If the propensity score is known ($s^m=0$), then only consistency for $\hat g_0$ is needed. Such comparisons also extend to the approximately sparse models. \qed \end{remark}

\begin{theorem}
 Suppose that  the ATE, $\theta_0 = \mathrm{E}_P[g_0(1,Z) - g_0(0,Z)]$,  is the target  parameter and we use the estimator $\tilde \theta_0$ and other notations defined above. Alternatively, suppose that the ATTE, $\theta_0 = \mathrm{E}_P[g_0(1,Z) -g_0(0,Z) \mid D=1]$, is the target parameter and we use the estimator $\tilde  \theta_0$ and other notations  above.  Consider the set $\mathcal{P}$ of probability distributions $P$ defined in Assumption \ref{ass:ATE}. Then,  
 uniformly in  $P \in \mathcal{P}$, the estimator $\tilde \theta_0$ concentrates
 around $\theta_0$ with the rate $1/\sqrt{N}$ and is approximately unbiased and normally distributed:
\begin{align*}
\sigma^{-1} \sqrt{N} (\tilde \theta_0- \theta_0) \rightsquigarrow  N(0,  1), \\
\sigma^2 = \mathrm{E}_P[\psi^2(W; \theta_0,\eta_0(Z))],
\end{align*}
and the result continues to hold if $\sigma^2$ is replaced by $\hat \sigma^2$.     Moreover, confidence regions based upon $\tilde \theta_0$ have uniform asymptotic validity:
\begin{align*}
& \sup_{P \in \mathcal{P}}| P \left ( \theta_0 \in \text{CI}_{n }\right) - (1- \alpha) | \to 0.
\end{align*}
The scores $\psi$ are the efficient scores, so both estimators are asymptotically
efficient, in the sense of reaching the semi-parametric efficiency bound of \citeasnoun{hahn:prop}.
\end{theorem} 

The proof follows from the application of Chebyshev's inequality and the central limit theorem (and is given in the appendix for completeness).

\section{Accounting for Uncertainty Due to Sample-Splitting}

The method outlined in this note relies on subsampling to form auxiliary samples for estimating nuisance functions and main samples for estimating the parameter of interest. The specific sample partition has no impact on estimation results asymptotically but may be important in finite samples. Specifically, the dependence of the estimator on the particular split creates an additional source of variation. Incorporating a measure of this additional source of variation into estimated standard errors of parameters of interest may be important for quantifying the true uncertainty of the parameter estimates. 

Hence we suggest making a slight modification to the asymptotically valid estimation procedure detailed in Section \ref{Sec: Algorithm}.  Specifically, we propose repeating the main estimation procedure $S$ times, for a large number $S$, repartitioning the data in each replication $s = 1,..., S$.  Within each partition, we then obtain an estimate of the parameter of interest, $\tilde\theta_0^s$.  Rather than report point estimates and interval estimates based on a single replication, we may then report estimates that incorporate information from the distribution of the individual estimates obtained from the $S$ different data partitions. 

For point estimation, two natural quantities that could be reported are the sample average
and the sample median  of the estimates obtained across the $S$ replications, $\tilde{\theta}_0^{\text{Mean}}$ and  $\tilde{\theta}_0^{\text{Median}}$.  Both of these reduce the sensitivity of the estimate for $\theta_0$ to particular splits.  $\tilde\theta_0^{\text{Mean}}$ could be strongly affected by any extreme point estimates obtained in the different random partitions of the data, and $\tilde\theta_0^{\text{Median}}$ is obviously much more robust.  We note that asymptotically the specific random partition is irrelevant, and $\tilde\theta_0^{\text{Mean}}$ and $\tilde\theta_0^{\text{Median}}$ should be close to each other.

To quantify and incorporate the variation introduced by sample splitting, one might also compute standard errors that add an element to capture the spread of the estimates obtained across the $S$ different sets of partitions.  For $\tilde\theta_0^{\text{Mean}}$, we propose adding an element that captures the spread of the estimated $\tilde\theta_0^s$ around $\tilde\theta_0^{\text{Mean}}$.  Specifically, we suggest
\begin{equation*}
\hat \sigma^{\text{Mean}} =\sqrt{ \frac{1}{S} \sum_{s=1}^{S} \left( \hat \sigma_{s}^2 + (\tilde \theta_0^{s} - \frac{1}{S} \sum_{j=1}^{S} \tilde \theta_{0}^j)^{2} \right) },
\end{equation*}
where $\hat \sigma_{s}$ is defined as in Section \ref{Sec: Algorithm}. The second term in this formula takes into  account the variation due to sample splitting which is added to a usual estimate of sampling uncertainty.  Using this estimated standard error obviously results in more conservative inference than relying on the $\hat\sigma_s$ alone.  We adopt a similar formulation for $\tilde\theta_0^{\text{Median}}$.  Specifically, we propose a median deviation defined as
\begin{align*}
\hat\sigma^{\text{Median}} = \underset{}{\text{median}}\Big\{ &\sqrt{ \hat \sigma_{i}^2 + (\hat \theta_{i} - \hat \theta^{Median})^{2} }  \Big\}_{i=1}^S.
\end{align*}
This standard error is more robust to outliers than $\hat\sigma^{\text{Mean}}$.

\appendix

\section{Practical Implementation and Empirical Examples}

To illustrate the methods developed in this paper, we consider two empirical examples. In the first, we use the method outlined in the paper to estimate the effect of 401(k) eligibility on accumulated assets. In this example, the treatment variable is not randomly assigned and we aim to eliminate the potential biases due to the lack of random assignment by flexibly controlling for a rich set of variables. The second example reexamines the Pennsylvania Reemployment Bonus experiment which used a randomized control trial to investigate the incentive effect of unemployment insurance. Our goal in this supplement is to illustrate the use of our method and examine its empirical properties in two different settings: 1) an observational study where it is important to flexibly control for a large number of variables in order to overcome endogeneity, and 2) a randomized control trial where controlling for confounding factors is not needed for bias reduction but may produce more precise estimates.

\subsection*{Practical Details: Incorporating Uncertainty Induced by Sample Splitting} The results we report are based on repeated application of the method developed in the main paper as discussed in Section III.  Specifically, we repeat the main  estimation procedure 100 times repartitioning the data in each replication.  We then report the average of the ATE estimates from the 100 random splits as the  ``Mean ATE," and we report the median of the ATE estimates from the 100 splits as the ``Median ATE."  We then report the measures of uncertainty that account for sampling variability and variability across the splits, $\hat \sigma^{\text{Mean ATE}}$ and $\hat \sigma^{\text{Median ATE}}$, for the ``Mean ATE'' and the ``Median ATE'' respectively.

\subsection*{The effect of 401(k) Eligibility on Net Financial Assets}  The key problem in determining the effect of 401(k) eligibility is that working for a firm that offers access to a 401(k) plan is not randomly assigned.  To overcome the lack of random assignment, we follow the strategy developed in \citen{pvw:94} and \citen{pvw:95}.  In these papers, the authors use data from the 1991 Survey of Income and Program Participation and argue that eligibility for enrolling in a 401(k) plan in this data can be taken as exogenous after conditioning on a few observables of which the most important for their argument is income.  The basic idea of their argument is that, at least around the time 401(k)'s initially became available, people were unlikely to be basing their employment decisions on whether an employer offered a 401(k) but would instead focus on income and other aspects of the job.  Following this argument, whether one is eligible for a 401(k) may then be taken as exogenous after appropriately conditioning on income and other control variables related to job choice.

A key component of the argument underlying the exogeneity of 401(k) eligibility is that eligibility may only be taken as exogenous after conditioning on income and other variables related to job choice that may correlate with whether a firm offers a 401(k).  \citen{pvw:94} and \citen{pvw:95} and many subsequent papers adopt this argument but control only linearly for a small number of terms.  One might wonder whether such specifications are able to adequately control for income and other related confounds.  At the same time, the power to learn about treatment effects decreases as one allows more flexible models.  The principled use of flexible machine learning tools offers one resolution to this tension.  The results presented below thus complement previous results which rely on the assumption that confounding effects can adequately be controlled for by a small number of variables chosen \textit{ex ante} by the researcher.

In the example in this paper, we use the same data as in \citen{CH401k}.  We use net financial assets - defined as the sum of IRA balances, 401(k) balances, checking accounts, U.S. saving bonds, other interest-earning accounts in banks and other financial institutions, other interest-earning assets (such as bonds held personally), stocks, and mutual funds less non-mortgage debt - as the outcome variable, $Y$, in our analysis.  Our treatment variable, $D$, is an indicator for being eligible to enroll in a 401(k) plan.  The vector of raw covariates, $Z$, consists of age, income, family size, years of education, a married indicator, a two-earner status indicator, a defined benefit pension status indicator, an IRA participation indicator, and a home ownership indicator.  

In Table \ref{table: 401k_mean}, we report estimates of the mean average treatment effect (Mean ATE) of 401(k) eligibility on net financial assets both in the partially linear model and allowing for heterogeneous treatment effects using the interactive model outlined in Section discussed in the main text.  To reduce the disproportionate impact of extreme propensity score weights in the interactive model we trim the propensity scores which are close to the bounds, with the cutoff points of 0.01 and 0.99.  We present two sets of results  based on sample-splitting  using a 2-fold cross-fitting and 5-fold cross-fitting.  

We report results based on five simple methods for estimating the nuisance functions used in forming the orthogonal estimating equations.  We consider three tree-based methods, labeled ``Random Forest'', ``Reg. Tree'', and ``Boosting'',  one $\ell_1$-penalization based method, labeled ``Lasso'', and a neural network method, labeled ``Neural Net".  For ``Reg. Tree,'' we fit a single CART tree to estimate each nuisance function with penalty parameter chosen by 10-fold cross-validation. The results in the ``Random Forest''  column are obtained by estimating each nuisance function with a random forest which averages over 1000 trees. The results in ``Boosting'' are obtained using boosted regression trees with regularization parameters chosen by 10-fold cross-validation. To estimate the nuisance functions using the neural networks, we use 8 hidden layers and a decay parameter of 0.01, and we set activation function as logistic for classification problems and as linear for regression problems.\footnote{We also experimented with ``Deep Learning''   methods from which we obtained similar results for some tuning parameters. However, we ran into stability and computational issues and chose not to report these results in the empirical section.} ``Lasso" estimates an $\ell_1$-penalized linear regression model using the data-driven penalty parameter selection rule developed in \citen{BellChenChernHans:nonGauss}. For ``Lasso'', we use a set of 275 potential control variables formed from the raw set of covariates and all second order terms, i.e. all squares and first-order interactions.  For the remaining methods, we use the raw set of covariates as features.

We also consider two hybrid methods labeled ``Ensemble" and ``Best".  ``Ensemble" optimally combines four of the machine learning methods listed above by estimating the nuisance functions as weighted averages of estimates from ``Lasso,'' ``Boosting,'' ``Random Forest,'' and ``Neural Net''.  The weights are restricted to sum to one and are chosen so that the weighted average of these methods gives the lowest average mean squared out-of-sample prediction error estimated using 5-fold cross-validation. The final column in Table \ref{table: 401k_mean} (``Best'') reports results that combines the methods in a different way.  After obtaining estimates from the five simple methods and ``Ensemble'',  we select the best methods for estimating each nuisance functions based on the average out-of-sample prediction performance for the target variable associated to each nuisance function obtained from each of the previously described approaches.  As a result, the reported estimate in the last column uses different machine learning methods to estimate different nuisance functions. Note that if a single method outperformed all the others in terms of prediction accuracy for all nuisance functions, the estimate in the ``Best" column would be identical to the estimate reported under that method.

Turning to the results, it is first worth noting that the estimated ATE of 401(k) eligibility on net financial assets is \$19,559 (not reported) with an estimated standard error of 1413 when no control variables are used.  Of course, this number is not a valid estimate of the causal effect of 401(k) eligibility on financial assets if there are neglected confounding variables as suggested by \citen{pvw:94} and \citen{pvw:95}.  When we turn to the estimates that flexibly account for confounding reported in Table \ref{table: 401k_mean}, we see that they are substantially attenuated relative to this baseline that does not account for confounding, suggesting much smaller causal effects of 401(k) eligiblity on financial asset holdings.  It is interesting and reassuring that the results obtained from the different flexible methods are broadly consistent with each other.  This similarity is consistent with the theory that suggests that results obtained through the use of orthogonal estimating equations and any sensible method of estimating the necessary nuisance functions should be similar.  Finally, it is interesting that these results are also broadly consistent with those reported in the original work of \citen{pvw:94} and \citen{pvw:95} which used a simple intuitively motivated functional form, suggesting that this intuitive choice was sufficiently flexible to capture much of the confounding variation in this example.

There are other interesting observations that can provide useful insights into understanding the finite sample properties of the ``double ML" estimation method. First, the standard errors of the estimates obtained using  5-fold cross-fitting are considerably lower than those obtained from 2-fold cross-fitting for all methods. This fact suggests that having more observations in the auxiliary sample may be desirable.  Specifically, 5-fold cross-fitting estimates uses more observations to learn the nuisance functions than 2-fold cross-fitting and thus likely learns them more precisely.  This increase in precision in learning the nuisance functions may then translate into more precisely estimated parameters of interest.  While intuitive, we note that this statement does not seem to be generalizable in that there does not appear to be a general relationship between the number of folds in cross-fitting and the precision of the estimate of the parameter of interest. Second, we also see that the standard errors of the Lasso estimates are noticeably larger than the standard errors coming from the other machine learning methods. We believe that this is due to the fact that the out-of-sample prediction errors from a linear model tend to be larger when there is a need to extrapolate. In our framework, if the main sample includes observations that are outside of the range of the observations in the auxiliary sample, the model has to extrapolate to those observations. The fact that  the standard errors are lower in 5-fold cross-fitting than in 2-fold cross-fitting for  the ``Lasso" estimations also supports this hypothesis, because the higher number of observations in the auxiliary sample reduces the degree of extrapolation.
 
In Table \ref{table: 401k_median}, we report the median ATE estimation results for the same models to check the robustness of our results to the outliers due to sample splitting. We see that both the coefficients and standard errors are similar to the ``Mean ATE"  estimates and standard errors. The similarity between the ``Mean ATE" and ``Median ATE" suggests that the distribution of the ATE across different splits is approximately symmetric and relatively thin-tailed.

\subsection*{The effect of Unemployment Insurance Bonus on Unemployment Duration}  As a further example, we re-analyze the Pennsylvania Reemployment Bonus experiment which was conducted by the US Department of Labor in the 1980s to test the incentive effects of alternative compensation schemes for unemployment insurance (UI). This experiment has been previously studied by \citen{B:PennSeq} and \citen{BK:PennQuan}. In these experiments, UI claimants were randomly assigned either to a control group or one of five treatment groups.\footnote{There are six treatment groups in the experiments. Following  \citen{B:PennSeq}. we merge the groups 4 and 6.} In the control group the standard rules of the UI applied. Individuals in the treatment groups were offered a cash bonus if they found a job within some pre-specified period of time (qualification period), provided that the job was retained for a specified duration. The treatments differed in the level of the bonus, the length of the qualification period, and whether the bonus was declining over time in the qualification period; see \citen{BK:PennQuan} for further details on data.

In our empirical example, we focus only on the most generous compensation scheme, treatment 4, and drop all individuals who received other treatments. In this treatment, the bonus amount is high and qualification period is long compared to other treatments, and claimants are eligible to enroll in a workshop. Our treatment variable, D, is an indicator variable for the treatment 4, and the outcome variable, Y, is the log of duration of unemployment for the UI claimants.  The vector of covariates, Z, consists of age group dummies, gender, race, number of dependents, quarter of the experiment, location within the state, existence of recall expectations, and type of occupation. 

We report estimates of the ATE on unemployment duration both in the partially linear model and in the interactive model. We again consider the same methods with the same tuning choices, with one exception, for estimating the nuisance functions as in the previous example and so do not repeat details for brevity. The one exception is that we implement neural networks with 2 hidden layers and a decay parameter of 0.02 in this example which yields better prediction performance.  In ``Lasso" estimation,  we use a set of 96 control variables formed by taking nonlinear functions and interactions of the raw set of covariates.  For the remaining approaches, we use only the 14 raw control variables listed above.


Table~\ref{table:  bonus_mean} presents estimates of the ``Mean ATE" on unemployment duration using the partially linear model and interactive model in panel A and B, respectively. To reduce the disproportionate impact of extreme propensity score weights in the interactive model, we trim the propensity scores at 0.01 and 0.99 as in the previous example.  For both the partially linear model and the interactive model, we report  estimates obtained using 2-fold cross-fitting and 5-fold cross-fitting.

The estimation results are consistent with the findings of previous studies which have analyzed the Pennsylvania Bonus Experiment. The ATE on unemployment duration is negative and significant across all estimation methods at the 5\% level with the exception of the estimate of the ATE obtained from the interactive model using random forests, which is significant at the 10\% level. When looking at standard errors  it is useful to remember that they include both sampling variation and variation due to random sample splitting. It is reassuring to see that the variation due to sample splitting does not change the conclusion. It is also interesting to see that, similar to the result in the first empirical example, the ``Mean ATE" estimates are broadly similar across different estimation models. Finally in Table~\ref{table: bonus_median} we report the ``Median ATE" estimates. The median estimates are close to the mean estimates, giving further evidence for the stability of estimation across different random splits.

In conclusion, we want to emphasize some important observations that can be drawn from these empirical examples. First, for both examples the choice of the machine learning method in estimating nuisance functions does not substantively change the conclusion, and we obtained broadly consistent results regardless of which method we employ. Second, the similarity between the median and mean estimates suggests that the results are robust to the particular sample split used in estimation in these examples.

\section{Proofs}

\subsection*{Notation} The symbols $\Pr$ and $\Ep$ denote probability and expectation operators with respect to a generic probability measure. If we need to signify the dependence on a probability measure  $P$, we use $P$ as a subscript in $\Pr_P$ and $\Ep_P$. Note also that we use capital letters such as $W$ to denote random elements and use the corresponding lower case letters such as $w$ to denote fixed values that these random elements can take in the set $\mathcal{W}$.  In what follows, we use $\|\cdot\|_{P,q}$ to denote the $L^q(P)$ norm; for example, for measurable $f: \mathcal{W} \to \Bbb{R}$, we denote $$\|f(W)\|_{P,q} := \left(\int |f(w)|^q d P(w)\right)^{1/q}.$$ 
Define the empirical process $\mathbb{G}_{n,I}(\psi(W))$ as a linear operator acting on measurable functions $\psi: \mathcal{W} \to \Bbb{R}$ such that $\|\psi(W)\|_{P,2}< \infty$ via,
$$
\mathbb{G}_{n,I}(\psi(W)) :=  \frac{1}{\sqrt{n}} \sum_{i \in I} \left( f(W_i) - \int f(w) dP(w)\right).
$$
Analogously, we define the empirical expectation and probability  as:
$$
 \mathbb{E}_{n,I}(\psi(W)) :=  \frac{1}{n} \sum_{i \in I}  f(W_i);   \quad
  \mathbb{P}_{n,I}(A) := \frac{1}{n} \sum_{i \in I}  1 ( W_i \in A).
$$

\subsection*{Proof of Theorem 1}


We will demonstrate the result for the case of ATE estimator,  which uses the score:
$$
\begin{array}{ll}
\psi(W; \theta,  \eta) := g(1,Z) - g(0,Z) + \frac{D (Y -g(1,Z)) } {m(Z)}  - \frac{(1-D) (Y -g(0,Z))}{1-m(Z)}  - \theta,
\end{array}
$$
and the result for ATTE follows similarly.
Choose any sequence $\{P_n\} \in \mathcal{P}$.

\noindent
{\bf Step 1:} (Main Step).    Letting $\check \theta_{0,k}  = \check \theta_0 (I_k, I_k^c)$,
write 
$$\sqrt{n} (\check \theta_k - \theta_0) = \mathbb{G}_{n,I_k} \psi(W; \theta_0, \hat \eta_0(I^c_k))
+ \sqrt{n} \int  \psi(w; \theta_0, \hat \eta_0(I^c_k)) d P_n(w).
$$
Steps 2 -- 5 below demonstrate that for each $k=1,...,K$,
\begin{eqnarray}
 && \int (\psi(w; \theta_0, \hat \eta_0(I^c_k))-\psi(w; \theta_0, \eta_0 ))^2 d P_n(w)  = o_{P_n}(1),
 \label{claim:1}\\
 &&  \sqrt{n} \int \psi(w; \theta_0, \hat \eta_0(I^c_k)) d P_n(w)  = o_{P_n}(1),  \label{claim:2} \\
 && \hat \sigma^2 - \sigma^2  = o_{P_n}(1),\label{claim:3}\\
&&\sigma^{-2} = O(1),\quad\text{and}\quad \sigma^2 = O(1).\label{claim:4}
\end{eqnarray}
The equations \eqref{claim:1} and \eqref{claim:2} are the minimal conditions needed on the estimators of the nuisance parameters, and could be used to replace the more primitive conditions stated in the text.

Assertion (\ref{claim:1}) implies that
$$
 \mathbb{G}_{n,I_k} ( \psi(W; \theta_0, \hat \eta_0(I^c_k)) - \psi(W; \theta_0,  \eta_0) ) =o_{P_n}(1),
$$
since the quantity in the display converges in probability conditionally 
on the data $(W_i)_{i \in I^c_k}$ by  (\ref{claim:1})  and Chebyshev inequality, which in turn implies the unconditional convergence in probability, as noted in the following simple lemma.

\begin{lemma}\label{conditional}Let $\{X_m\}$ and $\{Y_m\}$ be sequences of 
random vectors. If  for any $\epsilon > 0$, $\Pr (\|X_m\| > \epsilon \mid Y_m) \to_{\Pr} 0$,
then $\Pr(\|X_m\| > \epsilon ) \to 0$.  In particular, this occurs if  $\Ep [\|X_m\|^q \mid Y_m] \to_{\Pr} 0$ for some $q \geq 1$, by Markov's inequality. \end{lemma}

\noindent
{\bf Proof}. For any $\epsilon>0$ $\Pr(\|X_m\| > \epsilon) \leq \Ep [ \Pr(\|X_m\| > \epsilon \mid Y_m ) ]    \to 0$,  since the sequence $\{ \Pr(\|X_m\| > \epsilon \mid Y_m )\}$ is uniformly integrable. \qed

Using independence of data blocks  $(W_i)_{i \in I_k}$, $k =1, ..., K$, the application of the Lindeberg-Feller theorem and the Cramer-Wold device, we conclude that 
$$
\left ( \sigma^{-1} \sqrt{n} ( \check \theta_{0,k} - \theta_0 ) \right)_{k=1}^K  = \left (  \sigma^{-1} \mathbb{G}_{n,I_k} \psi(W; \theta_0,  \eta_0) \right)_{k=1}^K +o_{P_n}(1)  \leadsto   (\mathcal{N}_k)_{k=1}^K,
$$
where $(\mathcal{N}_k)_{k=1}^K$ is a Gaussian vector with independent $N(0,1)$ coordinates.
Therefore,
\begin{eqnarray*}
\sigma^{-1}\sqrt{nK} ( \tilde \theta_0 - \theta_0) & = &   \sigma^{-1} \sqrt{nK} \left(  \frac{1}{K} \sum_{k=1}^K\check \theta_{0,k} - \theta_0 \right) \\
 & =  &  \frac{1}{\sqrt{K}} \sum_{k=1}^K\sigma^{-1} \mathbb{G}_{n,I_k} \psi(W; \theta_0,  \eta_0) + o_{P_n}(1)
 \leadsto \frac{1}{\sqrt{K}} \sum_{k=1}^K \mathcal{N}_k =  N(0,1),
\end{eqnarray*}
where the last line uses the sum-stability of the normal distribution.  Moreover,
the result continues to hold if $\sigma$ is replaced by $\hat \sigma$ in view of (\ref{claim:3}) and \eqref{claim:4}.

The above claim implies that $\mathrm{CI}_n = [ \tilde \theta_0 \pm \Phi^{-1}(1-\alpha/2)\hat \sigma/\sqrt{N}]$ obeys
$$
\Pr_{P_n} (  \theta_0 \in \mathrm{CI}_n) \to (1-\alpha).
$$
In addition, the last two claims  hold under any sequence $\{P_n\} \in \mathcal{P}$, which implies that 
these claims hold uniformly in $P \in \mathcal{P}$.  Indeed, for example, choose $\{P_n\}$ such that,  for some $\epsilon_n \to 0$
$$
\sup_{P \in \mathcal{P}} |\Pr_P (  \theta_0 \in \mathrm{CI}_n) - (1-\alpha) |  
\leq |\Pr_{P_n} (  \theta_0 \in \mathrm{CI}_n) - (1-\alpha) |  + \epsilon_n.
$$
The right-hand side converges to zero, which implies the uniform convergence. It remains to prove claims \eqref{claim:1} -- \eqref{claim:4}.

\noindent
\textbf{Step 2:} This step demonstrates assertion (\ref{claim:1}). Observe that for some constant $C_{\varepsilon}$ that depends only on $\varepsilon$ and $\mathcal P$,
$$
\|  \psi(W; \theta_0, \hat \eta_0(I^c_k))-\psi(W; \theta_0; \eta_0 ) \|_{P_n,2} \leq C_{\varepsilon} (\mathcal I_1 + \mathcal I_2 + \mathcal I_3),
$$
where
\begin{align*}
&\mathcal I_1 = \max_{d \in \{0,1\}} \Big\| \hat g_0(d,Z; I^c_k) - g_0(d,Z) \Big\|_{P_n,2},\\
&\mathcal I_2 = \Big\| \frac{D(Y - \hat g_0(1,Z;I_k^c))}{\hat m_0(Z; I_k^c)} - \frac{D(Y - g_0(1,Z))}{m_0(Z)} \Big\|_{P_n,2},\\
&\mathcal I_3 = \Big\| \frac{(1 - D)(Y - \hat g_0(0,Z; I_k^c))}{1 - \hat m_0(Z; I_k^c)} - \frac{ (1 - D)(Y - g_0(0,Z)) }{1 - m_0(Z;I_k^c)} \Big\|_{P_n,2}.
\end{align*}
We bound $\mathcal I_1$, $\mathcal I_2$, and $\mathcal I_3$ in turn. First, $P_n(\mathcal I_1 > \delta_n) \leq \Delta_n \to 0 $ by Assumption \ref{ass:ATE}, and so $\mathcal I_1 = o_{P_n}(1)$. Also, on the event that 
\begin{equation}\label{eq: event 1}
P_n(\varepsilon \leq \hat m_0(Z;I^c_k) \leq 1- \varepsilon) =1
\end{equation}
and
\begin{equation}\label{eq: event 2}
\| \hat g_0(1,Z; I^c_k) - g_0(1,Z) \|_{P_n,2} +  \| \hat m_0(Z;I^c_k) - m_0(Z) \|_{P_n,2} \leq \delta_n,
\end{equation}
which happens with $\Pr_{P_n}$-probability at least $1 - \Delta_n$ by Assumption \ref{ass:ATE},
\begin{align*}
\mathcal I_2 
&\leq \varepsilon^{-2}\Big\| D m_0(Z)(Y - \hat g_0(1,Z;I_k^c)) - D\hat m_0(Z; I_k^c)(Y - g_0(1,Z)) \Big\|_{P_n,2}\\
&\leq \varepsilon^{-2}\Big\| m_0(Z)(g_0(1,Z) + \zeta - \hat g_0(1,Z;I_k^c)) - \hat m_0(Z; I_k^c)\zeta \Big\|_{P_n,2}\\
&\leq \varepsilon^{-2}\Big(\Big\| m_0(Z)(\hat g_0(1,Z;I_k^c) - g_0(1,Z)) \Big\|_{P_n,2} + \Big\| (\hat m_0(Z; I_k^c) - m_0(Z)) \zeta \Big\|_{P_n,2}\Big)\\
&\leq \varepsilon^{-2}\Big(\| \hat g_0(1,Z;I_k^c) - g_0(1,Z) \|_{P_n,2} + \sqrt C \|\hat m_0(Z; I_k^c) - m_0(Z)\|_{P_n,2}\Big) \leq \varepsilon^{-2}(\delta_n + \sqrt C \delta_n)\to 0,
\end{align*} 
where the first inequality follows from \eqref{eq: event 1} and Assumption \ref{ass:ATE}, the second from the facts that $D\in\{0,1\}$ and for $D = 1$, $Y = g_0(1,Z) + \zeta$, the third from the triangle inequality, the fourth from the facts that $P_n(m_0(Z) \leq 1) = 1$ and $P_n(\Ep_{P_n}[\zeta^2\mid Z]\leq C) = 1$, which are imposed in Assumption \ref{ass:ATE}, the fifth from \eqref{eq: event 2}, and the last assertion follows since $\delta_n \to 0$. Hence, $\mathcal I_2 = o_{P_n}(1)$. In addition, the same argument shows that $\mathcal I_3 = o_{P_n}(1)$, and so the claim of this step follows.

\noindent
\textbf{Step 3:}  This step demonstrates the assertion (\ref{claim:2}). Observe that since $\theta_0 = \Ep_{P_n}[g_0(1,Z) - g_0(0,Z)]$, the left-hand side of (\ref{claim:2}) is equal to  
\begin{align*}
\mathcal I_4&=\sqrt{n}  \int    \Bigg ( \frac{\hat m_0(z; I_k^c) - m_0(z)}{\hat m_0(z; I_k^c)}\cdot (\hat g_0(1,z; I^c_k) -  g_0(1,z))\\
& \quad +  \frac{\hat m_0(z; I_k^c) - m(z)}{1-\hat m_0(z; I_k^c)} \cdot (\hat g_0(0,z; I^c_k) - { g_0(0,z)} ) \Bigg ) d P_n(z).
\end{align*}
But on the event that 
$$
P_n(\varepsilon \leq \hat m_0(Z;I^c_k) \leq 1- \varepsilon) =1
$$
and
$$
 \max_{d\in\{0,1\}}  \| \hat g_0(d,Z; I^c_k) - g_0(d,Z) \|_{P_n,2} \cdot  \| \hat m_0(Z;I^c_k) - m_0(Z) \|_{P_n,2} \leq \delta_n n^{-1/2},
$$
which happens with $\Pr_{P_n}$-probability at least $1 - \Delta_n$ by Assumption \ref{ass:ATE}, we have using the Cauchy-Schwarz inequality that
$$
\mathcal I_4 \leq  \frac{2\sqrt n}{\varepsilon} \max_{d\in\{0,1\}}  \| \hat g_0(d,Z; I^c_k) - g_0(d,Z) \|_{P_n,2} \cdot  \| \hat m_0(Z;I^c_k) - m_0(Z) \|_{P_n,2}  \leq 
\frac{2 \delta_n}{\varepsilon}  \to 0,
$$
which gives the claim of this step.

\noindent
\textbf{Step 4:}   This step demonstrates the assertion (\ref{claim:3}). We can write
$$
\hat \sigma^2  = \frac{1}{K} \sum_{k=1}^K \hat \sigma^2_k, \quad \hat \sigma^2_k := \mathbb E_{n,I_k} \psi^2(W; \tilde \theta_k, \hat \eta_0(I^c_k)); \quad \sigma^2 = \Ep_{P_n} \psi^2(W; \theta_0, \eta_0).
$$
We claim that for each $k =1,..., K$,
$$
 \mathbb E_{n,I_k} \psi^2(W; \tilde \theta_0, \hat \eta_0 (I^c)) - \mathbb E_{n,I_k} \psi^2(W; \theta_0,  \eta_0) = o_{P_n}(1), \quad \mathbb{E}_{n,I_k} \psi^2(W; \theta_0,  \eta_0)  - \sigma^2 = o_{P_n}(1).
$$
The latter property holds by the Chebyshev Inequality. Further, letting $I$ denote a generic $I_k$,   the relation $a^2-b^2=(a-b)(a+b)$, and the Cauchy-Schwarz and triangle inequalities yield
 $$
| \mathbb E_{n,I} \{\psi^2(W; \hat \theta_0, \hat \eta_0 (I^c)) -  \psi^2(W; \theta_0,  \eta_0) \}|
\leq r_n  
\times \left( 2 \|\psi(W; \theta_0,  \eta_0)  \|_{\mathbb{P}_{n,I},2}  + r_n \right)
 $$
 where 
 $$
r_n := \| \psi(W; \tilde \theta_0,  \hat \eta_0( I^c)) -  \psi(W; \theta_0,  \eta_0)  \|_{\mathbb{P}_{n,I},2}.
$$
Hence, given that  $\|\psi(W; \theta_0,  \eta_0)  \|^2_{\mathbb{P}_{n,I},2} - \sigma^2 = o_{P_n}(1)$
as noted above, and $\sigma^2$ is bounded above by \eqref{claim:4}, the claim follows as long as we can show that 
\begin{equation}\label{eq: rn convergence}
r_n = o_{P_n}(1).
\end{equation}
To show \eqref{eq: rn convergence}, we have
$$
r_n \leq \|\tilde\theta_0 - \theta_0\| + \|  \psi(W; \theta_0, \hat \eta_0(I^c))-\psi(W; \theta_0; \eta_0 ) \|_{\mathbb P_{n,I},2},
$$
and $\|\tilde \theta_0 - \theta_0\| = o_{P_n}(1)$ by the arguments in Step 1. Also,
\begin{align*}
&\Ep_{P_n}\Big[\|  \psi(W; \theta_0, \hat \eta_0(I^c))-\psi(W; \theta_0; \eta_0 ) \|_{\mathbb P_{n,I},2} \mid (W_i)_{I\in I^c}\Big]\\
&\qquad \leq \|  \psi(W; \theta_0, \hat \eta_0(I^c_k))-\psi(W; \theta_0; \eta_0 ) \|_{P_n,2} = o_{P_n}(1).
\end{align*}
by Jensen's inequality and Step 2. Hence, \eqref{eq: rn convergence} holds, which gives the claim of this step.
%

\noindent
\textbf{Step 5:}   This step demonstrates the assertion (\ref{claim:4}). Note that the bound $\sigma^2 = O(1)$ holds trivially from Assumption \ref{ass:ATE}. Hence, it suffices to show that $\sigma^{-2} = O(1)$. To do so, observe that using Assumption \ref{ass:ATE} we have
\begin{align*}
\Ep\Big[\psi^2(W,\theta_0,\eta_0)\Big]
&=\Ep\Big[\Ep[\psi^2(W,\theta_0,\eta_0) \mid D,Z]\Big]\\
&=\Ep\Big[\Ep[(g_0(1,Z) - g_0(0,Z) - \theta_0)^2 \mid D,Z] \\
&\quad + \Ep\Big[\Big( \frac{D(Y - g_0(1,Z))}{m_0(Z)} - \frac{(1 - D)(Y - g_0(0,Z))}{1 - m_0(Z)} \Big)^2 \mid D,Z\Big]\Big]\\
&\geq \Ep\Big[\Big( \frac{D(Y - g_0(1,Z))}{m_0(Z)} - \frac{(1 - D)(Y - g_0(0,Z))}{1 - m_0(Z)} \Big)^2\Big]\\
& = \Ep\Big[\Big( \frac{(1 - m_0(Z))D(Y - g_0(1,Z)) - m_0(Z)(1 - D)(Y - g_0(0,Z))}{m_0(Z)(1 - m_0(Z))} \Big)^2\Big]\\
& = \Ep\Big[\Big( \frac{(D - m_0(Z))\zeta}{m_0(Z)(1 - m_0(Z))} \Big)^2\Big] \geq \Ep[(D - m_0(Z))^2 \zeta^2] = \Ep[\nu^2 \zeta^2] \geq \Ep[\nu^2]\Ep[\zeta^2]\geq c^4,
\end{align*}
where the last line follows since $P_n(m(Z) \leq 1) = P_n(1 - m(Z) \leq 1) = 1$. This gives the claim of this step and completes the proof of the theorem.\qed

\bibliography{MLPnPArxiv} 

\bibliographystyle{econometrica}

\clearpage

\begin{table}[ht]
\footnotesize
\centering
\caption{Estimated Mean ATE of 401(k) Eligibility on Net Financial Assets}\label{table: 401k_mean}
\begin{tabular}{l@{\hskip 0.8 in}ccccccc}
  \hline  \hline  \\  [-0.1in] 
 & Lasso & Reg. Tree & Random Forest & Boosting & Neural Net. & Ensemble & Best \\ \\  [-0.20in]  \cline{2-8}  \\  [-0.20in]     \multicolumn{2}{l}{\textit{A. Interactive Model}} \\ \\   [-0.1in]
ATE (2 fold)    &6331 & 7581 & 7966 & 7826 & 7805 & 7617 & 7800 \\ 
       & (2712) & (1374) & (1549) & (1345) & (1688) & (1299) & (1325)  \\   
ATE (5 fold)    &6964 & 8023 & 8104 & 7699 & 7772 & 7658 & 7890 \\ 
      &  (1654) & (1311) & (1364) & (1223) & (1324) & (1204) & (1198)  \\   \\ [-0.1in]
   \multicolumn{2}{l}{\textit{B. Partially Linear Model}} \\ \\ [-0.1in]
ATE (2 fold)    & 7718 & 8745 & 9180 & 8768 & 9040 & 9043 & 9106 \\ 
       &  (1796) & (1488) & (1526) & (1451) & (1494) & (1432) & (1430)  \\  
ATE (5 fold)    & 8182 & 8913 & 9248 & 9092 & 9038 & 9186 & 9214 \\ 
      & (1578) & (1440) & (1402) & (1380) & (1394) & (1381) & (1361)  \\   \\ [-0.15in]
   \hline \hline \\[-0.3cm]
     \multicolumn{8}{l}{%
  \begin{minipage}{17cm}%
    \footnotesize Notes: Estimated Mean ATE and standard errors (in parentheses) from a linear model (Panel B) and heterogeneous effect model (Panel A) based on orthogonal estimating equations. Column labels denote the method used to estimate nuisance functions. Further details about the methods are provided in the main text.  \\
  \end{minipage}%
}\\
\end{tabular}
\end{table}

\begin{table}[ht]
\footnotesize
\centering
\caption{Estimated Median  ATE of 401(k) Eligibility on Net Financial Assets}\label{table: 401k_median}
\begin{tabular}{l@{\hskip 0.8 in}ccccccc}
  \hline  \hline  \\  [-0.1in] 
 & Lasso & Reg. Tree & Random Forest & Boosting & Neural Net. & Ensemble & Best \\ \\  [-0.20in]  \cline{2-8}  \\  [-0.20in]     \multicolumn{2}{l}{\textit{A. Interactive Model}} \\  \\   [-0.1in]
ATE (2 fold)    &    6725 & 7557 & 8034 & 7820 & 7800 & 7620 & 7800 \\ 
       &  (1612) & (1283) & (1400) & (1199) & (1474) & (1198) & (1185)  \\  
ATE (5 fold)    & 7133 & 8046 & 8099 & 7690 & 7795 & 7668 & 7876 \\ 
      & (1420) & (1242) & (1296) & (1179) & (1290) & (1180) & (1149)  \\   \\   [-0.1in]
   \multicolumn{2}{l}{\textit{B. Partially Linear Model}} \\ \\ [-0.1in]
ATE (2 fold)    & 7707 & 8770 & 9204 & 8746 & 9104 & 9061 & 9129  \\ 
       & (1785) & (1424) & (1392) & (1391) & (1388) & (1343) & (1342)   \\  
ATE (5 fold)    & 8202 & 8894 & 9252 & 9089 & 9065 & 9199 & 9232 \\ 
      & (1581) & (1440) & (1400) & (1378) & (1393) & (1379) & (1359)  \\   \\  [-0.15in]
   \hline \hline \\[-0.3cm]
     \multicolumn{8}{l}{%
  \begin{minipage}{17cm}%
    \footnotesize Notes: Estimated Median ATE and standard errors (in parentheses) from a linear model (Panel B) and heterogeneous effect model (Panel A) based on orthogonal estimating equations. Column labels denote the method used to estimate nuisance functions. Further details about the methods are provided in the main text.  \\
  \end{minipage}%
}\\
\end{tabular}
\end{table}

\clearpage

\begin{table}[ht]
\footnotesize
\centering
\caption{Estimated Mean ATE of Cash Bonus on Unemployment Duration}\label{table: bonus_mean}
\begin{tabular}{l@{\hskip 0.8 in}ccccccc}
  \hline  \hline  \\  [-0.1in] 
 & Lasso & Reg. Tree & Random Forest & Boosting & Neural Net. & Ensemble & Best \\ \\  [-0.20in]  \cline{2-8}  \\  [-0.20in]  
   \multicolumn{2}{l}{\textit{A. Interactive Model}} \\ \\   [-0.1in]
ATE (2 fold)    & -0.081 & -0.084 & -0.072 & -0.078 & -0.073 & -0.079 & -0.078  \\ 
       &  (0.036) &  (0.037) & (0.042) & (0.036) & (0.041) & (0.036) & (0.036) \\   
ATE (5 fold)    & -0.081 & -0.084 & -0.070 & -0.076 & -0.072 & -0.079 & -0.076  \\ 
      &  (0.036) & (0.037) & (0.040) & (0.036) & (0.038) & (0.036) & (0.036) \\   \\   [-0.1in]
   \multicolumn{2}{l}{\textit{B. Partially Linear Model}} \\ \\ [-0.1in]
ATE (2 fold)    & -0.081 & -0.083 & -0.076 & -0.076 & -0.073 & -0.076 & -0.076 \\ 
       &  (0.036) & (0.037) & (0.037) & (0.036) & (0.036) & (0.036) & (0.036)  \\  
ATE (5 fold)    &  -0.080 & -0.084 & -0.075 & -0.075 & -0.074 & -0.075 & -0.075   \\ 
      & (0.036) & (0.037) & (0.036) & (0.036) & (0.036) & (0.036) & (0.036)   \\   \\  [-0.15in]
   \hline \hline \\[-0.3cm]
     \multicolumn{8}{l}{%
  \begin{minipage}{17cm}%
    \footnotesize Notes: Estimated Mean ATE and standard errors (in parentheses) from a linear model (Panel B) and heterogeneous effect model (Panel A) based on orthogonal estimating equations. Column labels denote the method used to estimate nuisance functions. Further details about the methods are provided in the main text.  \\
  \end{minipage}%
}\\
\end{tabular}
\end{table}

\begin{table}[ht]
\footnotesize
\centering
\caption{Estimated Median ATE of Cash Bonus on Unemployment Duration}\label{table: bonus_median}
\begin{tabular}{l@{\hskip 0.8 in}ccccccc}
  \hline  \hline  \\  [-0.1in] 
 & Lasso & Reg. Tree & Random Forest & Boosting & Neural Net. & Ensemble & Best \\ \\  [-0.20in]  \cline{2-8}  \\  [-0.20in]  
   \multicolumn{2}{l}{\textit{A. Interactive Model}} \\ \\   [-0.1in]
ATE (2 fold)    & -0.081 & -0.084 & -0.073 & -0.078 & -0.074 & -0.079 & -0.078 \\ 
       & (0.036) & (0.036) & (0.041) & (0.036) & (0.039) & (0.036) & (0.036)  \\   
ATE (5 fold)    & -0.081 & -0.085 & -0.069 & -0.076 & -0.072 & -0.079 & -0.076\\ 
      &   (0.036) & (0.037) & (0.039) & (0.036) & (0.038) & (0.036) & (0.036) \\   \\    [-0.1in]
   \multicolumn{2}{l}{\textit{B. Partially Linear Model}} \\ \\ [-0.1in]
ATE (2 fold)    &  -0.081 & -0.084 & -0.077 & -0.076 & -0.074 & -0.076 & -0.076 \\ 
       &  (0.036) & (0.036) & (0.036) & (0.036) & (0.036) & (0.036) & (0.036)  \\  
ATE (5 fold)    &  -0.079 & -0.084 & -0.076 & -0.075 & -0.073 & -0.075 & -0.075 \\ 
      & (0.036) & (0.037) & (0.036) & (0.035) & (0.036) & (0.035) & (0.035) \\   \\  [-0.15in]
   \hline \hline \\[-0.3cm]
     \multicolumn{8}{l}{%
  \begin{minipage}{17cm}%
    \footnotesize Notes: Estimated Median ATE and standard errors (in parentheses) from a linear model (Panel B) and heterogeneous effect model (Panel A) based on orthogonal estimating equations. Column labels denote the method used to estimate nuisance functions. Further details about the methods are provided in the main text.  \\
  \end{minipage}%
}\\
\end{tabular}
\end{table}

\end{document}